\pdfoutput=1
\documentclass[11pt]{article}

\usepackage[preprint]{acl}

\usepackage{times}
\usepackage{latexsym}

\usepackage[T1]{fontenc}
\usepackage[utf8]{inputenc}

\usepackage{microtype}

\usepackage{inconsolata}

\usepackage{graphicx}

\usepackage{booktabs}
\usepackage{multirow}
\usepackage{xcolor}
\usepackage{colortbl}
\usepackage{arydshln}
\usepackage{tcolorbox}
\usepackage{listings}
\usepackage{amsmath}
\usepackage{tabularx}

\title{Tag-Evol: Achieving Efficient Instruction Evolving via Tag Injection}

\author{
 \textbf{Yixuan Wang\footnotemark[2]},
 \textbf{Shiqi Zhou\footnotemark[2]},
 \textbf{Chuanzhe Guo},
 \textbf{Qingfu Zhu\footnotemark[1]}
\\
Research Center for Social Computing and Interactive Robotics,
\\
Harbin Institute of Technology, China
\\
\texttt{\{wyx,sqzhou,czguo,qfzhu\}@ir.hit.edu.cn}
}

\begin{document}\maketitle

\renewcommand{\thefootnote}{\fnsymbol{footnote}}
% \footnotetext[2]{The two authors contribute equally to this work.}
\footnotetext[2]{Equal contribution.}
\footnotetext[1]{Corresponding author.}
\renewcommand{\thefootnote}{\arabic{footnote}}

\begin{abstract}
% Background
Evol-Instruct has made significant improvements as a data synthesis method in several areas.
% Promblems
Existing methods typically rely on a fixed set of strategies to evolve, which require manual design and are monolithic in form.
In addition, iterative evolution also makes the acquisition of hard samples expensive.
% We introduce
In view of this, we propose the Tag-Evol framework, a more diverse and efficient instruction evolving method.
% Specific
Specifically, Tag-Evol uses diverse and specific knowledge tags as strategies to achieve controlled evolution by injecting different combinations of tags into the original instructions.
% Tag method
% Moreover, we propose a multi-step fine-grained tagging approach to provide richer knowledge tags for evolutions.
% Experiment
Experiments with multiple backbones in diverse domain benchmarks show that the proposed method generates significantly better evolved data than other methods.
% Besides
Furthermore, we conduct a thorough analysis of the evolved data, demonstrating that Tag-Evol is not only efficient but also generates more diverse and challenging data.

\end{abstract}

\section{Introduction}

% LLM -> SFT -> Synthetic Data
As a key stage in the training of large language models (LLMs), supervised fine-tuning (SFT) greatly improves the performance of the models on various natural language processing tasks \citep{ouyang2022training,wang2023aligning}.
To achieve better alignment results in the SFT phase, many high-quality instruction datasets (e.g., ShareGPT \citep{chiang2023vicuna}, LIMA \citep{zhou2024lima}) have been collected and proposed.
However, human-annotated high-quality datasets are usually expensive and limited, which cannot support larger scale SFT of the model.

% Self-Instruct -> Evol and Auto Evol
In light of this, recent researches \citep{tan2024large,dubey2024llama,yang2024qwen2} has focused on how to synthesize instruction data for training using LLMs themselves, which is relatively cheap and scalable.
Evol-Instruct \citep{xu2023wizardlm} achieves high-quality instruction synthesis by utilizing rewriting ability of LLMs.
This method iteratively improves the difficulty and diversity of the samples from an initial seed dataset through heuristic prompts, which are ultimately merged as the instruction dataset.
Benefiting from well-designed evolution prompts and the power of LLMs, Evol-Instruct has achieved a wide range of applications in several domains \citep{luo2023wizardmath,luo2023wizardcoder}.
% TODO Add Auto-Evol
Besides, \citet{zeng2024automatic} leverage LLMs to automatically optimize the evolution prompts, which further improves the quality of the evolution process.

\begin{figure}[t]
  \includegraphics[width=\columnwidth]{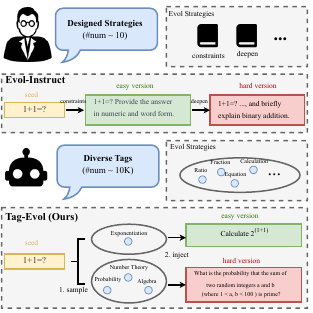}
  \caption{An illustration of the differences between the Evol-Instrcut and Tag-Evol.
  Tag-Evol uses diverse tags as evolution strategies and can generate difficult sample at once by controlling the number of tags.
  }
  \label{fig:intro}
\end{figure}

% Weakness?
While Evol-Instruct has shown excellent results on a variety of tasks and models, there are still some issues that need to be addressed:
(1) \textbf{Fixed Evolution Strategies}.
Existing methods obtain a fixed set of evolution strategies through heuristic human annotation and machine optimization, which is costly and hardly adaptable to other domains.
Simultaneously, fixed strategies also limit the diversity of evolved data due to preference bias of the model itself.
(2) \textbf{Inefficient Synthesis Mode}.
In order to obtain the hard version of instructions, Evol-Instruct requires multiple iterations of evolution, which not only take extra time, but also introduce cumulative errors due to hallucinations.
An ideal instruction evolving algorithm should have diverse and specific evolution strategies and be able to efficiently generate synthetic data of varying difficulty.

% What we do ?
To address the above challenges, we propose \textbf{Tag-Evol} framework, 
a more efficient approach for evolved instruction data synthesis through knowledge tag information injection.
Inspired by \citet{lu2023instag}, who evaluates the difficulty of instructions based on the number of knowledge tags typed by LLMs,
we redesign the instruction evolving method from the perspective of knowledge tags.
These tags represent the attributes contained in the instruction.
As shown in Figure \ref{fig:intro}, we use diverse and specific knowledge tags as enriched evolution strategies, and use different combinations and numbers of knowledge tags to explicitly control the degree of evolution.
In this way we can avoid the fixed strategy by combining tags and control the number of injected tags to directly generate hard samples without iteration.

Specifically, our Tag-Evol framework can be divided into two parts, tag pool construction and tag sampling evolution.
% TagPool
Firstly, we obtain a domain-specific tag pool by tagging and merging the samples in the seed dataset.
Unlike tagging methods used for evaluation, we want to acquire tags that are as diverse and specific as possible, and for this reason we propose a multi-step fine-grained knowledge tagging approach (see details in \S \ref{TagPhrase}) that can significantly improve the quantity and quality of the tag pool.
% Evol
After obtaining the tag pool, we guide the synthesis of the instruction data by sampling different numbers of knowledge tags as the evolution strategy (see details in \S \ref{EvolPhrase}).
By prompting the LLMs to inject the knowledge of different tag combinations into the seed dataset, we are able to achieve more accurate instruction evolution.
% Based on the findings in InsTag \cite{lu2023instag} that the difficulty of instruction depends on the num of knowledge tags it contains, we achieve a degree of difficulty-controlled data evolution by controlling the number of injected tags.

% Experiments
% We have conducted experiments using several mainstream LLMs as backbones on both mathematical reasoning and code generation tasks.
We have conducted experiments using several mainstream LLMs as backbones on multiple tasks including multi-turn conversation, instruction following, mathematical reasoning, and code generation.
For a fair comparison, we replicat the Evol-Instrut \citep{xu2023wizardlm} and Auto Evol-Instruct \citep{zeng2024automatic} as our baseline methods under the same open source setup.
The results show that the proposed Tag-Evol significantly outperforms the baseline model under different backbones, achieving better downstream task performance on multiple domains.
In addition, we have also conducted exhaustive analytical experiments on the proposed method in terms of the size of the required evolution model, the number of iterative convergence, demonstrating the great potential of the Tag-Evol method in the field of synthetic data.
Our main contributions can be summarised as follows:
\begin{itemize}
    \item We propose the Tag-Evol framework\footnote{\url{https://github.com/fghccv/TagEvol}}, which achieves efficient instruction evolving by injecting different combinations of knowledge tags.
    \item We propose a multi-step fine-grained knowledge tagging approach that can obtain more diverse and specific tags for evolution.
    \item Experiments show that the proposed method outperforms existing Evol-Instruct methods on diverse tasks under multiple base models.
\end{itemize}
% Contribution

% synthetic Data Influence For LLM, Using LLM 

% Systheic Data Evol

% Systhetic Date (self-instruct, evol-instruction)

% Evol Instruction disadventage

\begin{figure*}[ht]
  \includegraphics[width=\textwidth]{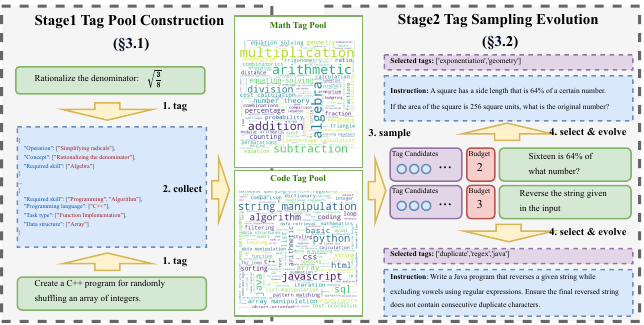}
  \caption{An illustration of the proposed Tag-Evol framework.
First, we utilize LLMs to tag seed datasets to construct diverse and fine-grained tag pools.
In the evolution phase, we sample a batch of candidate tags and let the model select the appropriate tags to inject into the original samples according to the budget, achieving an increase in difficulty.
}
  \label{fig:main}
\end{figure*}

\section{Related Work}

\paragraph{Synthesis with LLMs.}
Compared to human annotated datasets, synthetic data has a wide range of applications \citep{tan2024large,dubey2024llama,yang2024qwen2,abdin2024phi} in synthetic instruction datasets due to its lower cost.
\citet{wang2022self} first propose the Self-Instruct method, which utilizes the LLM itself to efficiently synthesize instruction data for alignment.
Subsequently, synthetic data (e.g., Alpaca \citep{taori2023alpaca}, CodeAlpaca \citep{codealpaca}, etc.) based on Self-Instruct targeting various domains are released for training in the SFT phase.
The Evol-Instruct proposed by the Wizard series \citep{xu2023wizardlm,luo2023wizardmath,luo2023wizardcoder} dramatically increases the upper bound of the available instruction data through well-designed evolution strategies.
To address the high cost of manually designed strategies, \citet{zeng2024automatic} propose an automated strategy optimization method to achieve a wider range of Evol-Instruct.

While most synthetic data methods are implemented using closed-source models (e.g., ChatGPT \citep{achiam2023gpt}), there has been some recent works focus on how to use open-source models for data synthesis.
\citet{xu2024magpie} exploit the design of open-source model templates and utilize special tokens to distill high quality instruction data from the SFT phase of the model.
\citet{ding2024unleashing} further propose the question fine-tuning (QFT) task, which takes instruction data synthesis as a task for the model to learn.
It enables the fine-tuned QFT model to extract more high-quality instruction data.
\citet{hui2024smaller} compare the performance of different scale models in the Evol-Instruct method and demonstrate that small models are more suitable for generating diverse evolved data.

\paragraph{Evaluation with LLMs.}
In addition to the generation of synthetic data, how to automatically evaluate and filter synthetic data using LLMs is also an important issue.
\citet{zheng2024judging} propose the MT-Bench dataset to evaluate the performance of LLM-based chat assistants by scoring them with a powerful LLM judge.
\citet{liu2023makes} effectively evaluate and filter the SFT phase training data in terms of three dimensions: complexity, quality, and diversity.
Unlike directly scoring, \citet{lu2023instag} propose the InsTag method to assess the difficulty and diversity of instruction data from the perspective of knowledge tags. 
By tagging each sample using LLMs, InsTag can estimate the difficulty of different samples by the number of tags and the diversity of the dataset by the total number of tags.

InsTag proposes a systematic approach to data evaluation, but existing synthetic data approaches do not take full advantage of the benefits of tags in knowledge representation.
In this paper, we propose the Tag-Evol framework for efficient and high-quality instruction evolution using the content and number of tags as evolution strategies.

\section{Tag-Evol Framework}

As shown in Figure \ref{fig:main}, our Tag-Evol framework can be divided into two parts as a whole: tag pool construction (\S\ref{TagPhrase}) and tag sampling evolution (\S\ref{EvolPhrase}).
In the tag pool construction phase, our goal is to extract tags containing extensive knowledge from the seed dataset as strategies for subsequent evolution.
After obtaining the tags, we implement the evolution of instructions with various levels of difficulty by randomly sampling a certain number of tags and injecting them into the original dataset utilizing the rewriting ability of LLMs.

\subsection{Tag Pool Construction}
\label{TagPhrase}
% InsTag and Promblem
InsTag \citep{lu2023instag} implement a straightforward tagging method that aims to obtain coarse-grained tags, which can be used to estimate the difficulty and diversity of the dataset.
However, for evolution strategies, coarse-grained tags do not provide rich and specific guidance goals.
To this end, we propose a multi-step fine-grained tagging method that is capable of yielding a larger number of more specific tags for subsequent use.

\begin{figure}[t]
  \includegraphics[width=\columnwidth]{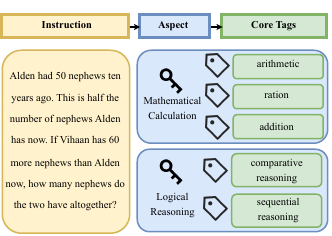}
  \caption{An illustration of the proposed multi-step fine-grained tagging method.}
  \label{fig:Tag}
\end{figure}

% Our method Step1 & Step2
Specifically, we specify the structure of the original tags from lists to dictionaries, allowing the model to generate more comprehensive knowledge tags from keys to values step by step.
As shown in Figure \ref{fig:Tag}, the whole process is divided into two parts:
(1) \textbf{Aspect Generation}.
For a more comprehensive tagging, we first prompt the model to summarize in abstract terms, describing the macroscopic characteristics of the sample.
For example, the type of task, the skills required, the type of arithmetic, and so on. These abstract aspects ensure the diversity of evolution strategies, with the ability to change the task categories of the samples, add new constraints.
(2) \textbf{Tag Generation}.
Subsequently, we ask the model to generate concrete knowledge tags based on the abstract aspects, which will be sampled in combination as final evolution strategies.
It is these specific tags that distinguish Tag-Evol from previous Evol-Instruct method.
Tag-Evol can give more specific details of constraints (e.g., exponential correlation) than a generalized strategy of adding constraints, thus making the goal of evolution more clear and the content of evolution more diverse.

The model generates both content together in chain of thought format (see Appendix \ref{app:prompt} for prompt details).
The analytical experiment in Section \ref{ab:tagway} demonstrate the advantages of our multi-step tagging over single-step, significantly improving the performance of Tag-Evol.

\subsection{Tag Sampling Evolution}
\label{EvolPhrase}
The motivation for tag sampling evolution is to inject samples with the proper tag combination so that they can contain more tags and thus evolve in difficulty.
% input output
Specifically, after obtaining the tag pool $\mathcal{P}$, for each sample $x$, we assign it a difficulty budget $b$ indicating the number of tags to be injected into it.

\paragraph{Budget controlled injection.}
In order to minimize the insertion of irrelevant tags that lead to hallucinations, we choose to provide the evolution model with a batch collection of tags, so that it can choose the appropriate subset for evolution according to the budget.
For the evolution model $M_\theta$, the entire tag injection process can be formulated as:
\begin{equation}
    \hat{x}, t = M_\theta(x, b, cand)
    \label{eq1}
\end{equation}
where $x$ represents the original data, $b$ is the difficulty budget, and $cand$ is a subset sampled from $\mathcal{P}$.
For the output of the model, $\hat{x}$ represents the evolved instruction, and $t$ is the tag combination chosen by $M_\theta$ with $t \subset cand$ and $\lvert t \rvert=b$.

Similar to Auto Evol-Instruct \citep{zeng2024automatic}, we use designed prompt to make $M_\theta$ complete the complex operations in Equation \ref{eq1}.
The whole tag injection process can be divided into four steps (see Appendix \ref{app:prompt} for details of prompt):
(1) Select proper tag combination from candidate tag batch based on the instruction and budget.
(2) Generate an injection plan based on original instruction and selected tags.
(3) Execute the plan to complete the generation of evolution instruction.
(4) Rewrite the instruction to remove the hallucinations.
The proposed process allows the model to choose a more appropriate combination of tags based on different instructions.
Together with the budget hyper-parameter, it can provide a unique evolution strategy for each sample.
In addition, we have conducted experiments for the scale of the evolution model $M_\theta$ in Section \ref{ab:scale}.
Due to the provision of specific evolution strategies in the form of tags, we find that even 7B level models can execute the Tag-Evol well.

\paragraph{Evolution in multiple rounds.}
Although Tag-Evol can directly generate samples of varying difficulty by choosing different budgets,
we choose to perform multiple rounds of evolution on the seed dataset in order to make a fair comparison with the Evol-Instruct method.
Considering that the Evol-Instruct method usually performs three rounds of evolution and merge them as the final dataset,
we also used three different budgets to generate three rounds of synthesized data with different levels of difficulty.
% Specifically, we use the budgets of 1 tag, 3 tags and 5 tags for the math domain and 3 tags, 5 tags, 7 tags for the code domain due to its less informative tags.

\section{Experiment}

\subsection{Experiment Setup}

\paragraph{Dataset setup.}
% Math and Code domain
To evaluate the proposed Tag-Evol framework effectively,
we choose three commonly used domains for validation, general-purpose task, mathematical reasoning and code generation, all of which require the model to have strong reasoning capabilities.
% benchmark
% General
The multi-turn conversation dataset MT-Bench \citep{zheng2023judging} and the instruction-following evaluation benchmark IFEval \citep{zhou2023instruction} are selected to represent general-purpose tasks.
% Math & Code
For the math domain, we use the GSM8K \citep{cobbe2021training} and the MATH-500 \citep{lightman2023let} as benchmarks, while for the code domain, we use the HumanEval \citep{chen2021evaluating} and MBPP \citep{austin2021program} benchmarks.
% Seed Dataset
% For the seed dataset, we choose the corresponding training set of GSM8K, MATH for math domain.
% Besides, we follow the previous setup using CodeAlpaca \citep{codealpaca}.
We construct seed datasets across three domains:
for general-purpose tasks, we use the Databricks-Dolly dataset \citep{DatabricksBlog2023DollyV2};
for the math domain, the training sets of GSM8K and MATH are adopted;
and for the code domain, we follow prior work by incorporating CodeAlpaca \citep{codealpaca}.
% Iteration
Detailed statistical information is shown in Table \ref{tab:seed}.

\begin{table}[t]
    \centering
    \begin{tabular}{ccc}
    \toprule
    Domain & Source & Num\# \\
    \midrule
    General & Databricks-Dolly & 15,000\\
    Math & GSM8K, MATH  & 7,473 + 7,500 \\
     Code & Code Alpaca & 20,022 \\
    \bottomrule
    \end{tabular}
    \caption{Statistical information on seed datasets used.}
    \label{tab:seed}
\end{table}

\paragraph{Model details.}
% Evol Model
Existing datasets vary in their seed sources and the evolution models used, making comparisons difficult.
To fairly evaluate existing Evol-Instruct methods, we construct detailed experiments using Qwen2.5-72B-Instruct \citep{yang2024qwen2} as an evolution model.
% Output
The model is also used to generate all the responses of the intruction data.
% Back Bone Model
In addition, we choose to train on three different pre-trained models Mistral-7b-v0.1 \citep{jiang2023mistral}, Llama3-8B \citep{dubey2024llama} and Qwen2.5-7B \citep{yang2024qwen2} which represent backbones with different abilities.
% HyperParameter
See Appendix \ref{app:hyper} for specific training setups and hyper-parameter details.

\begin{table*}[t]
    \centering 
    \begin{tabularx}{\textwidth}{c*8{>{\centering\arraybackslash}X}}
    \toprule
    \multirow{2}{*}{\textbf{Method}} &  \multicolumn{3}{c}{\textbf{General-purpose}} &  \multicolumn{2}{c}{\textbf{Math Reasoning}} & \multicolumn{2}{c}{\textbf{Code Generation}} & \multirow{2}{*}{\textbf{Average}}\\
    \cmidrule(l){2-4}
    \cmidrule(l){5-6}
    \cmidrule(l){7-8}
    & MTBench & IF-L & IF-S & GSM & MATH & HEval & MBPP &  \\
    \midrule
    \rowcolor{gray!15} \multicolumn{9}{c}{\textit{Mistral-7B-base}} \\
    Seed & 5.4 & 31.6 & 27.3 & 58.1 & 25.0 & 39.6 & 50.5 & 33.9\\
    \hdashline
    Evol-Ins & 5.2 & 36.2 & 32.9 & 63.3 & 33.4 & 53.0 & 56.3 & 40.0 \\
    Auto Evol-Ins & 5.6 & 40.6 & 34.3 & 65.3 & \textbf{35.4} & 53.7 & 55.0 & 41.4 \\
    Tag-Evol \textbf{(Ours)} & \textbf{5.7} & \textbf{43.8} & \textbf{39.5} & \textbf{67.6} & \textbf{35.4} & \textbf{55.5} & \textbf{58.2} & \textbf{43.7}\\
    \rowcolor{gray!15} \multicolumn{9}{c}{\textit{Llama3-8B-base}} \\
    Seed & 6.0 & 40.4 & 35.4 & 68.4 & 36.6 & 45.7 & 59.0 & 41.6\\
    \hdashline
    Evol-Ins & 6.2 & 46.0 & 40.4 & 70.6 & 42.3 & 59.8 & \textbf{66.4} & 47.4\\
    Auto Evol-Ins & 6.6 & 48.9 & 44.7 & 70.7 & 41.0 & 58.5 & 65.3 & 48.0 \\
    Tag-Evol \textbf{(Ours)} & \textbf{6.8} & \textbf{51.2} & \textbf{47.1} & \textbf{72.6} & \textbf{46.2} & \textbf{62.8} & \textbf{66.4} & \textbf{50.4} \\
    \rowcolor{gray!15} \multicolumn{9}{c}{\textit{Qwen2.5-7B-base}} \\
    Seed & 6.7 & 48.4 & 44.1 & 87.4 & 68.2 & 78.0 & 79.9 & 59.0\\
    \hdashline
    Evol-Ins & 6.8 & 48.4 & 44.5 & 87.1 & 69.0 & 78.7 & 78.8 & 59.0\\
    Auto Evol-Ins & 6.9 & 52.3 & 46.9 & 88.4 & 69.2 & 78.7 & \textbf{80.2} & 60.4 \\
    Tag-Evol \textbf{(Ours)} & \textbf{7.2} & \textbf{54.0} & \textbf{49.0} & \textbf{89.4} & \textbf{71.6} & \textbf{80.5} & \textbf{80.2} & \textbf{61.7} \\
    \bottomrule
    \end{tabularx}
    \caption{The main experimental results cover the domains of general-purpose tasks, mathematical reasoning, and code generation.
IF-L and IF-S denote the IFeval-loose and IFeval-strictly evaluation benchmarks, respectively.
}
    \label{tab:main}
\end{table*}

\paragraph{Baseline methods.}
In this paper, we choose two representative Evol-Instruct methods as baselines:
(1) \textbf{Evol-Instruct} \citep{xu2023wizardlm} manually design a number of evolution prompt as strategies, and complete each round of evolution by randomly selecting different strategies.
Specifically, we follow \citeposs{hui2024smaller} setup \footnote{https://github.com/HypherX/Evolution-Analysis/} and use four depth instructions and one width instruction as evolutionary strategies.
(2) \textbf{Auto Evol-Instruct} \citep{zeng2024automatic} leverage LLMs to reflect on generated evolutionary trajectories in order to optimize the original evolution strategy automatically, enabling it to generate higher quality data.
In our experiments, we directly choose the optimal prompt reported in Auto Evol-Instruct as the evolution strategy, which has been automatically optimized 12 times.

% iteration setting
In our experiments on downstream tasks, we adapt the setup from \citet{zeng2024automatic} and conduct three rounds of evolution merging to establish the baseline dataset.
% ours
For the proposed Tag-Evol, we choose budgets of 1, 3, and 5 tags for math and 3, 5, and 7 tags for code to generate the same number of datasets for comparison, respectively.

\subsection{Evaluation Results}

\paragraph{Main results.} 
% describe the result
After constructing the evolved dataset, we conduct experiments on three different base models as backbones
% to evaluate their performance in the mathematical and the code domain.
to evaluate their performance across three domains: general-purpose, mathematical reasoning, and code generation.
The results are shown in Table \ref{tab:main}, Tag-Evol significantly outperforms the existing Evol-Instruct method at all backbone settings, delivering an improvement of 2-3 points on average.
% Backbone
For some models released early such as Mistral, whose overall performance is relatively low, the diverse and complex evolved data generated by Tag-Evol can well enhance its performance.
% Math
For more advanced models like Qwen2.5, the power of its own capabilities results in a decrease for improvements of all Evol-Instruct methods.
But this case Tag-Evol still gives a boost of about 1.5 points compared to other evolutionary methods.

% \paragraph{Results under different backbones.}
% backbone promblem
In particular, we also observe a trend in the performance difference between the evolved data and the seed data under different backbone models.
% mistral big
In both the Mistral and Llama3 settings, the evolutionary approach shows a 8-9 points improvement compared to the seed dataset,
suggesting that by further increasing the difficulty of the knowledge involved in the seed dataset can help to improve the model's capabilities.
% qwen small
When it comes to the Qwen2.5 setting, however, the performance of the existing evolved dataset is very close to that of the seed dataset,
despite the fact that they have three times as much data as the seed dataset.
We analyze that it may be that the Qwen2.5 model has already mastered the knowledge involved in the seed dataset, and the SFT phase serves more as alignment than learning new skills.
% weakness
Existing Evol-Instruct methods evolve only based on one single sample and do not bring sufficiently diverse new combinations to stimulate new knowledge in LLMs.
% strength
The proposed Tag-Evol can be seen as a weighted combination of different samples in the seed dataset by injecting knowledge tags, leading to a more diverse evolved samples, and also a 2-point improvement in the Qwen2.5 strong backbone setting.

\section{Analysis}

\subsection{Effect of Multi-step Tagging}
\label{ab:tagway}
\begin{figure}[t]
  \includegraphics[width=\linewidth]{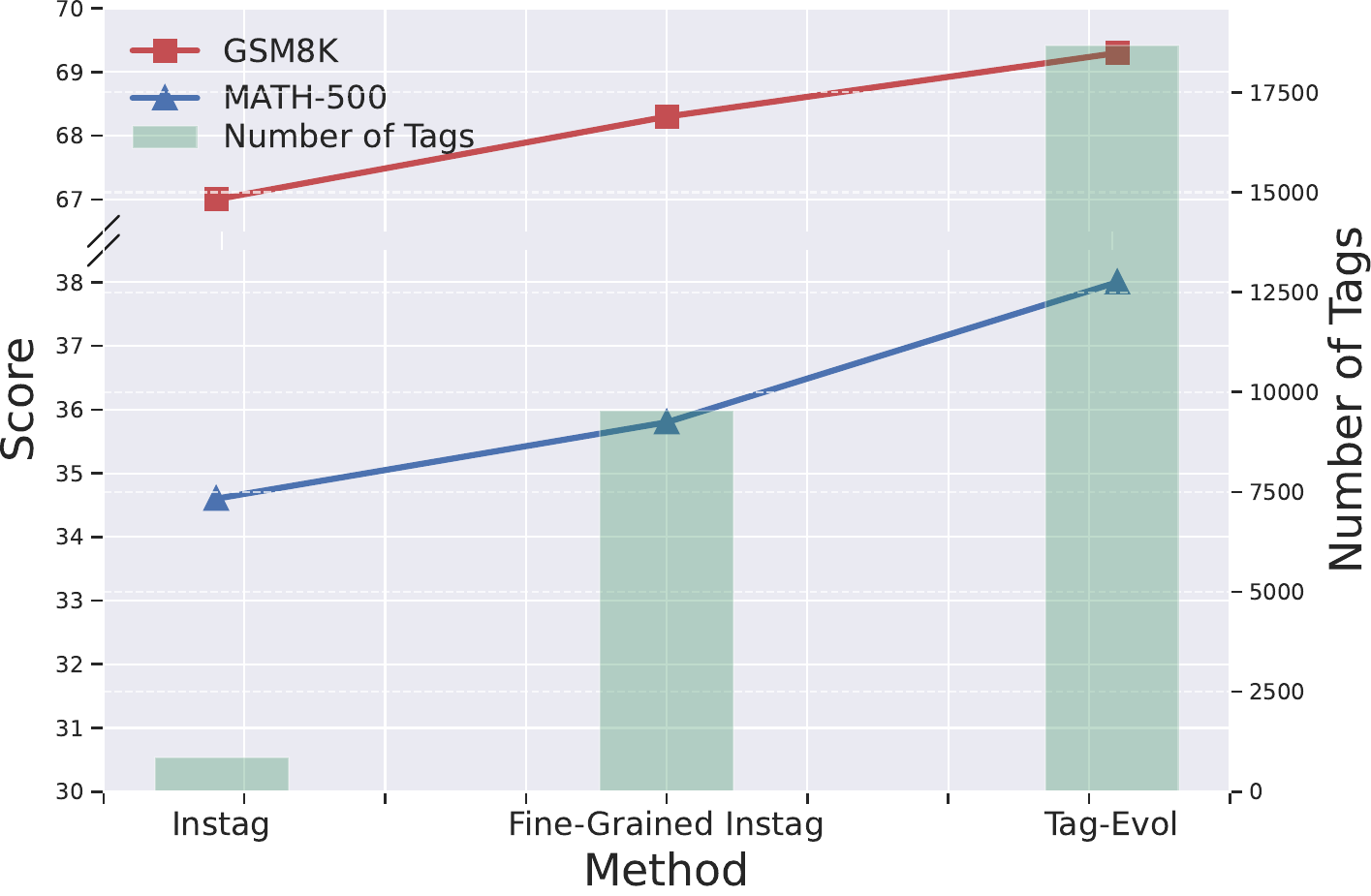}
  \caption{Impact of different tagging methods on the quality of the final evolved dataset.
  The y-axis represents task performance and the number of tag pools, respectively}
  \label{fig:tagway}
\end{figure}

\paragraph{Setup.} As mentioned in Section \ref{TagPhrase}, the diversity and specificity of tags directly affects the effectiveness of the Tag-Evol method.
In order to validate the effectiveness of the proposed fine-grained tagging method, we conduct analytical experiments on the tagging method.
% baseline
We choose two baseline methods: the original method reported in the \citeposs{lu2023instag} paper, and the method of modifying its content for one-step fine-grained tagging.
We compare the performance of Tag-Evol method under different tagging pools on the math domain setting using llama3-8B as the base model.
% result
\paragraph{Results.}
The experiment results are shown in Figure \ref{fig:tagway}.
% tag num
The most obvious difference between these methods is the number of tags in the tag pool, and with the two-phase tagging method, we can obtain tags about more aspects (20 times for the original, 2 times for the single step).
% performance
The number of tags represents the diversity of evolutionary strategies.
As shown in the table, the mathematical performance of the model increases gradually as the number of tags increases, from 67.0 to 69.3 on GSM8K.
For the more difficult MATH-500 dataset, the improvement from complex tags is even more significant, from 34.6 to 38.0.

\begin{figure*}[ht]
  \includegraphics[width=\textwidth]{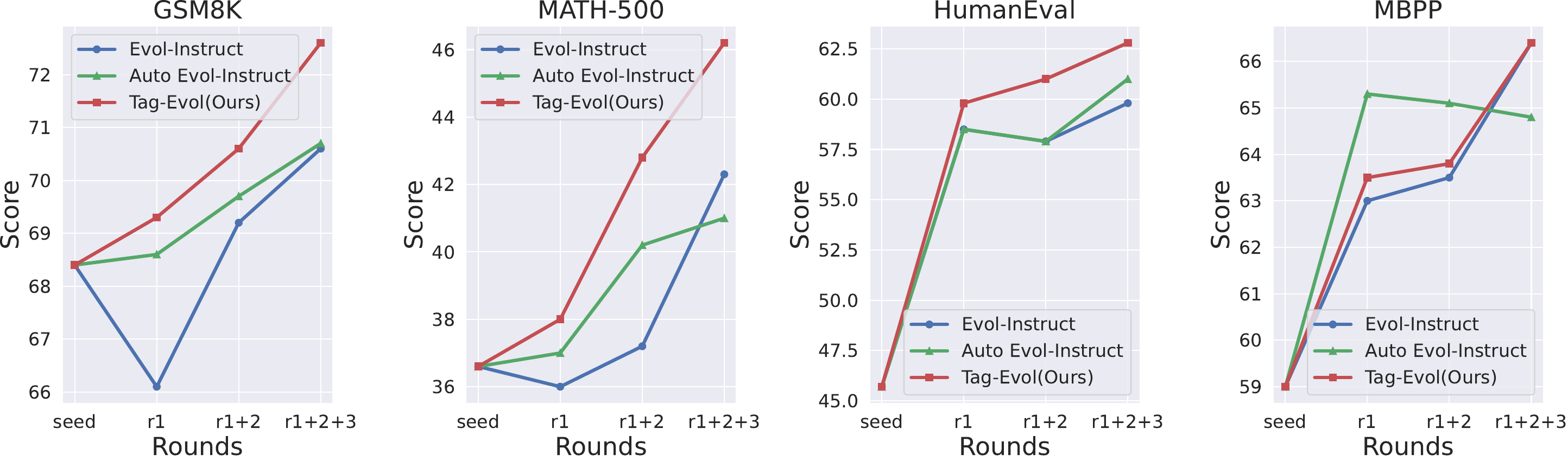}
  \caption{Multi-round performance at different data sizes using Llama3-8B as backbone.}
  \label{fig:datascale}
\end{figure*}

\subsection{Analysis of Data Scale}
\paragraph{Setup.}
In the main experiment, we use a 3-iteration evolutionary setup for all methods.
In this section we analyze the data at different scales for each iteration in more detail.
We compare the performance on Llama3-8B with medium capacity under three different amounts of evolved data.
It is worth noting that Evol-Instruct methods use iterative evolution, where the input to each round of evolution is the output of the previous round.
By contrast, all three rounds of the proposed Tag-Evol are evolved on the same seed dataset with different budgets.

\paragraph{Results.}
The performance of each method for different number of rounds is shown in Figure \ref{fig:datascale}.
% Scaling
As the number of evolution rounds increases, the size of the merged dataset keeps growing and all the evolutionary methods show an upward trend compared to the seed dataset.
% perfermance
Overall, Tag-Evol demonstrate strengths at almost all data scales across various benchmarks.
It suggests that injecting different numbers of knowledge tags directly to the original instructions generates equally or even more challenging samples compared to iterative evolution, resulting in improved performance for training on evolved data.

% more stable
In addition to this, Tag-Evol shows a more stable performance improvement compared to the Evol-Instruct method.
Our analysis suggests that this is due to the fact that strategy of Evol-Instruct is selected randomly, independent of the original instruction.
This may produce some inappropriate evolutions leading to hallucinations.
Differently, our approach mitigates this effectively by sampling a batch of tags for the model to actively select, enabling more stable training.

\subsection{Analysis of Model Scale}
\label{ab:scale}

\begin{table}[t]
    \centering
    \begin{tabular}{ccc}
        \toprule
        \multirow{2}{*}{\textbf{Method}} & \multicolumn{2}{c}{\textbf{Evolution LLM}} \\
        \cmidrule(l){2-3}
        & 7B-Instruct & 72B-Instruct \\
        \midrule
        \rowcolor{gray!15}  \multicolumn{3}{c}{\textit{Evol-Instruct}} \\
        Evol-Ins & \textbf{68.3} & 66.1 \\
        Auto Evol-Ins & 68.1 & \textbf{68.6} \\
        \midrule
        \rowcolor{gray!15}  \multicolumn{3}{c}{\textit{Tag-Evol}} \\
        7B-Tagger & \textbf{68.7} & 68.2 \\
        72B-Tagger & 69.2 & \textbf{69.3} \\
        \bottomrule
    \end{tabular}
    \caption{Analysis experiments on the scale of tagging and evolution models using the Qwen2.5 series of models.
    7B-Tagger represents the use of Qwen2.5-7B-Instruct to build the mentioned tag pool.}
    \label{tab:model_scale}
\end{table}

\begin{table*}[ht]
    \centering
    \begin{tabular}{ccccccccc}
        \toprule
        \multirow{2}{*}{\textbf{Method}} & \multicolumn{2}{c}{\textbf{GSM8K}} & \multicolumn{2}{c}{\textbf{MATH-500}} & \multicolumn{2}{c}{\textbf{HumanEval}}  & \multicolumn{2}{c}{\textbf{MBPP}}\\
        \cmidrule(l){2-3}
        \cmidrule(l){4-5}
        \cmidrule(l){6-7}
        \cmidrule(l){8-9}
        & 8-gram & 13-gram & 8-gram & 13-gram & 8-gram & 13-gram & 8-gram & 13-gram \\
        \midrule
        Seed & 60 & 4 & 528 & 76 & 38 & 0 & 83 & 2\\
        Evol-Ins & 31 & 1 & 341 & 31 & 33 & 0 & 46 & 2 \\
        Auto Evol-Ins & 38 & 0 & 388 & 24 & 26 & 0 & 5 & 1\\
        Tag-Evol & 30 & 3 & 357 & 36 & 15 & 0 & 50 & 1 \\
        \bottomrule
    \end{tabular}
    \caption{Leakage testing of different evolved datasets on benchmark data.}
    \label{tab:lake}
\end{table*}

\paragraph{Setup.}
% Motivcation
In order to evaluate in detail the requirements of the proposed method on the capability of instruct model in both the tag pool construction and tag sampling evolution phases, we conduct experiments using different model scales.
% Setting
We choose the Qwen2.5 series 7B-Instruct and 72B-Instruct as the evolutionary LLMs to evaluate the performance of three different Evol-Instruct approaches.
% Tagger
In addition to this, we also used these two scales of models to construct tag pools (7B-Tagger and 72B-Tagger in Tabel \ref{tab:model_scale}) separately to explore the relationship between the quality of tags and the ability of the tagging model.
% Output
Regardless of the way the instruction datasets are constructed, we use the Qwen2.5-72B-Instruct model to generate all the responses of the instruction data for a fair comparison.

\paragraph{Results.}
We conduct experiments on the GSM8K dataset and the results are shown in Table \ref{tab:model_scale}.
% for two Evol less important
For Evol-Instruct and Tag-Instruct, where the strategy has been given, we find that the evolved instructions generated by the smaller model achieve higher performance, consistent with \citeposs{hui2024smaller} findings.
It is shown that given an evolutionary strategy, evolution is essentially a rewriting task that does not require much modeling capability.
The uncertainty of small models may bring diversity to evolved data.
% for auto important
However, for the Auto Evol-Instruct approach, it requires the LLM to generate concrete strategies based on abstract prompt, resulting in its evolution being more dependent on the capabilities of the evolution model.

% Tag important
For the models used for tagging, the experiments illustrate that we still need some high-quality tags generated by large models to better guide the evolutionary process.
% Efficient
Fortunately, high-quality tags are not disposable and are easy to collect and maintain.
Once we have maintained a collection of high-quality tags in a specific domain, using Tag-Evol to evolve existing data will be an efficient process.
% As opposed to using expensive large models, we only need the rewriting ability of small models and a high-quality tag strategy to achieve evolution.

\subsection{Data Leakage Detection}
\paragraph{Setup.}
% For synthetic data, performing data leakage detection is an important part of the process.
We follow the \citeposs{liu2023tinygsm} setup and use gram level matching to detect data leakage.
Specifically, we choose 8-gram matching and 13-gram matching on four benchmarks in the math and code domains.

\paragraph{Results.}
As shown in Table \ref{tab:lake}, all the evolved datasets have a smaller number of matches to the benchmark than the seed dataset, having a lower risk of data leakage.
The matching number in the math domain is significantly higher than that in the code domain, which we analyze may be due to the fact that the seed set uses a homogeneous training set in math.
In addition, the high number of matches in the MATH dataset may be because this dataset has more formulas, which are not tokenized efficiently and are prone to multi-gram matches.

\begin{table}[t]
    \centering
    \begin{tabular}{ccccc}
        \toprule
        \textbf{Method} & \textbf{Seed} & \textbf{R1} &\textbf{R2} & \textbf{R3} \\
        \midrule
        \rowcolor{gray!15} \multicolumn{5}{c}{\textit{Difficulty}} \\
        Evol-Ins       & 2.12    & 2.38    & 2.30    & 2.44    \\
        Auto Evol-Ins  & 2.12    & 2.32    & 2.76    & 2.89    \\
        Tag-Evol   & 2.12    & \textbf{2.44}    & \textbf{2.96}    & \textbf{3.28}    \\
        \midrule
        \rowcolor{gray!15} \multicolumn{5}{c}{\textit{Diversity}} \\
        Evol-Ins       & 45      & \textbf{61}      & 66      & 62      \\
        Auto Evol-Ins  & 45      & 51      & \textbf{69}      & 71      \\
        Tag-Evol   & 45      & 60      & \textbf{69}      & \textbf{103}     \\
        \bottomrule
    \end{tabular}
    \caption{Automated evaluation metrics defined by InsTag for evolved data, where R1 represents  Round 1.}
    \label{tab:instag}
\end{table}

\subsection{Discussion of Complexity and Diversity}
\paragraph{Setup.}
In addition to the performance on downstream tasks, we also use the automated metric used in InsTag to evaluate our evolved data,
which is divided into two dimensions: diversity and difficulty.
% define
\citet{lu2023instag} use the average number of tags to indicate the difficulty of the dataset, and the full set of tags in the overall dataset to indicate diversity.
We follow his setting and experiment using the prompt reported in the paper.
% we use api
Specifically, we use the latest GPT-4o (gpt-4o-2024-08-06) for our experiments.
We randomly select 50 samples and evaluate different versions of these data evolved different rounds using different methods.

\paragraph{Results.}
The results are shown in Table \ref{tab:instag}. The Tag-Evol evolved dataset has significant advantages in both difficulty and diversity.
% Difficulty
For the difficulty metric, the experiment shows that our multi-tag direct injection strategy is feasible.
Compared to Evol-Instruct methods that iteratively evolve 3 times, we achieve the goal of efficiently constructing difficult samples by injecting multiple knowledge tags into the original instructions at once.
% Diversity
For the diversity metric, the samples evolved by Tag-Evol contain significantly more tags than the other methods in the third round by injecting multiple tags.
It shows that the knowledge tags we inject via Tag-Evol are effective and can be validated by re-tagging the evolved data.

\section{Conclusion}
In this paper, we propose a novel instruction evolving method, TagEvol,
that can generate diverse and challenging instruction data more efficiently.
Experiments in multiple domains illustrate the advantages of the Tag-Evol framework.
Its efficient generation with flexible strategies makes it a very promising method for data synthesis.

\section*{Limitations}
% API
Considering the cost factor, the main experiments in this paper all use the current powerful open source model Qwen2.5-72B-Instruct for tag pool construction and tag sampling generation.
The Tag-Evol evolved dataset may have higher performance if experiments are conducted using more powerful closed-source models, allowing the trained model to be compared with existing open-source SOTA models.
% Tag in Domain
Moreover, for a fair comparison, only the tags extracted from the seed dataset are chosen as the evolutionary strategy in this paper.
Whether by mixing high-quality tags from different sources can bring unexpected enhancement in the general domain is a worthwhile thing to try, 
which will also serve as our future research direction.

\section*{Ethics Statement}
The source data for the proposed methods is derived entirely from publicly available project resources on legitimate websites, ensuring that no sensitive information is involved. Furthermore, all baselines and datasets utilized in our experiments are also publicly accessible, and we have credited the respective authors by citing their work.

\section*{Acknowledgements}
% We gratefully acknowledge the support of the National Natural Science Foundation of China (NSFC) via grant 62236004, 62206078, 62441603 and 62476073.
We gratefully acknowledge the support of the National Natural Science Foundation of China (NSFC) via grant 62236004, 62206078 and 62476073.

\bibliography{custom}

\begin{thebibliography}{31}
\providecommand{\natexlab}[1]{#1}

\bibitem[{Abdin et~al.(2024)Abdin, Aneja, Awadalla, Awadallah, Awan, Bach, Bahree, Bakhtiari, Bao, Behl et~al.}]{abdin2024phi}
Marah Abdin, Jyoti Aneja, Hany Awadalla, Ahmed Awadallah, Ammar~Ahmad Awan, Nguyen Bach, Amit Bahree, Arash Bakhtiari, Jianmin Bao, Harkirat Behl, et~al. 2024.
\newblock Phi-3 technical report: A highly capable language model locally on your phone.
\newblock \emph{arXiv preprint arXiv:2404.14219}.

\bibitem[{Achiam et~al.(2023)Achiam, Adler, Agarwal, Ahmad, Akkaya, Aleman, Almeida, Altenschmidt, Altman, Anadkat et~al.}]{achiam2023gpt}
Josh Achiam, Steven Adler, Sandhini Agarwal, Lama Ahmad, Ilge Akkaya, Florencia~Leoni Aleman, Diogo Almeida, Janko Altenschmidt, Sam Altman, Shyamal Anadkat, et~al. 2023.
\newblock Gpt-4 technical report.
\newblock \emph{arXiv preprint arXiv:2303.08774}.

\bibitem[{Austin et~al.(2021)Austin, Odena, Nye, Bosma, Michalewski, Dohan, Jiang, Cai, Terry, Le et~al.}]{austin2021program}
Jacob Austin, Augustus Odena, Maxwell Nye, Maarten Bosma, Henryk Michalewski, David Dohan, Ellen Jiang, Carrie Cai, Michael Terry, Quoc Le, et~al. 2021.
\newblock Program synthesis with large language models.
\newblock \emph{arXiv preprint arXiv:2108.07732}.

\bibitem[{Chaudhary(2023)}]{codealpaca}
Sahil Chaudhary. 2023.
\newblock Code alpaca: An instruction-following llama model for code generation.
\newblock \url{https://github.com/sahil280114/codealpaca}.

\bibitem[{Chen et~al.(2021)Chen, Tworek, Jun, Yuan, Pinto, Kaplan, Edwards, Burda, Joseph, Brockman et~al.}]{chen2021evaluating}
Mark Chen, Jerry Tworek, Heewoo Jun, Qiming Yuan, Henrique Ponde De~Oliveira Pinto, Jared Kaplan, Harri Edwards, Yuri Burda, Nicholas Joseph, Greg Brockman, et~al. 2021.
\newblock Evaluating large language models trained on code.
\newblock \emph{arXiv preprint arXiv:2107.03374}.

\bibitem[{Chiang et~al.(2023)Chiang, Li, Lin, Sheng, Wu, Zhang, Zheng, Zhuang, Zhuang, Gonzalez et~al.}]{chiang2023vicuna}
Wei-Lin Chiang, Zhuohan Li, Zi~Lin, Ying Sheng, Zhanghao Wu, Hao Zhang, Lianmin Zheng, Siyuan Zhuang, Yonghao Zhuang, Joseph~E Gonzalez, et~al. 2023.
\newblock Vicuna: An open-source chatbot impressing gpt-4 with 90\%* chatgpt quality.
\newblock \emph{See https://vicuna. lmsys. org (accessed 14 April 2023)}, 2(3):6.

\bibitem[{Cobbe et~al.(2021)Cobbe, Kosaraju, Bavarian, Chen, Jun, Kaiser, Plappert, Tworek, Hilton, Nakano et~al.}]{cobbe2021training}
Karl Cobbe, Vineet Kosaraju, Mohammad Bavarian, Mark Chen, Heewoo Jun, Lukasz Kaiser, Matthias Plappert, Jerry Tworek, Jacob Hilton, Reiichiro Nakano, et~al. 2021.
\newblock Training verifiers to solve math word problems.
\newblock \emph{arXiv preprint arXiv:2110.14168}.

\bibitem[{Conover et~al.(2023)Conover, Hayes, Mathur, Xie, Wan, Shah, Ghodsi, Wendell, Zaharia, and Xin}]{DatabricksBlog2023DollyV2}
Mike Conover, Matt Hayes, Ankit Mathur, Jianwei Xie, Jun Wan, Sam Shah, Ali Ghodsi, Patrick Wendell, Matei Zaharia, and Reynold Xin. 2023.
\newblock \href {https://www.databricks.com/blog/2023/04/12/dolly-first-open-commercially-viable-instruction-tuned-llm} {Free dolly: Introducing the world's first truly open instruction-tuned llm}.

\bibitem[{Ding et~al.(2024)Ding, Shi, Liang, Li, Zhu, and Zhang}]{ding2024unleashing}
Yuyang Ding, Xinyu Shi, Xiaobo Liang, Juntao Li, Qiaoming Zhu, and Min Zhang. 2024.
\newblock Unleashing reasoning capability of llms via scalable question synthesis from scratch.
\newblock \emph{arXiv preprint arXiv:2410.18693}.

\bibitem[{Dubey et~al.(2024)Dubey, Jauhri, Pandey, Kadian, Al-Dahle, Letman, Mathur, Schelten, Yang, Fan et~al.}]{dubey2024llama}
Abhimanyu Dubey, Abhinav Jauhri, Abhinav Pandey, Abhishek Kadian, Ahmad Al-Dahle, Aiesha Letman, Akhil Mathur, Alan Schelten, Amy Yang, Angela Fan, et~al. 2024.
\newblock The llama 3 herd of models.
\newblock \emph{arXiv preprint arXiv:2407.21783}.

\bibitem[{Hui et~al.(2024)Hui, Zhao, Dong, Zhang, Zhou, and Su}]{hui2024smaller}
Tingfeng Hui, Lulu Zhao, Guanting Dong, Yaqi Zhang, Hua Zhou, and Sen Su. 2024.
\newblock Smaller language models are better instruction evolvers.
\newblock \emph{arXiv preprint arXiv:2412.11231}.

\bibitem[{Jiang et~al.(2023)Jiang, Sablayrolles, Mensch, Bamford, Chaplot, Casas, Bressand, Lengyel, Lample, Saulnier et~al.}]{jiang2023mistral}
Albert~Q Jiang, Alexandre Sablayrolles, Arthur Mensch, Chris Bamford, Devendra~Singh Chaplot, Diego de~las Casas, Florian Bressand, Gianna Lengyel, Guillaume Lample, Lucile Saulnier, et~al. 2023.
\newblock Mistral 7b.
\newblock \emph{arXiv preprint arXiv:2310.06825}.

\bibitem[{Lightman et~al.(2023)Lightman, Kosaraju, Burda, Edwards, Baker, Lee, Leike, Schulman, Sutskever, and Cobbe}]{lightman2023let}
Hunter Lightman, Vineet Kosaraju, Yura Burda, Harri Edwards, Bowen Baker, Teddy Lee, Jan Leike, John Schulman, Ilya Sutskever, and Karl Cobbe. 2023.
\newblock Let's verify step by step.
\newblock \emph{arXiv preprint arXiv:2305.20050}.

\bibitem[{Liu et~al.(2023{\natexlab{a}})Liu, Bubeck, Eldan, Kulkarni, Li, Nguyen, Ward, and Zhang}]{liu2023tinygsm}
Bingbin Liu, Sebastien Bubeck, Ronen Eldan, Janardhan Kulkarni, Yuanzhi Li, Anh Nguyen, Rachel Ward, and Yi~Zhang. 2023{\natexlab{a}}.
\newblock Tinygsm: achieving> 80\% on gsm8k with small language models.
\newblock \emph{arXiv preprint arXiv:2312.09241}.

\bibitem[{Liu et~al.(2023{\natexlab{b}})Liu, Zeng, He, Jiang, and He}]{liu2023makes}
Wei Liu, Weihao Zeng, Keqing He, Yong Jiang, and Junxian He. 2023{\natexlab{b}}.
\newblock What makes good data for alignment? a comprehensive study of automatic data selection in instruction tuning.
\newblock \emph{arXiv preprint arXiv:2312.15685}.

\bibitem[{Lu et~al.(2023)Lu, Yuan, Yuan, Lin, Lin, Tan, Zhou, and Zhou}]{lu2023instag}
Keming Lu, Hongyi Yuan, Zheng Yuan, Runji Lin, Junyang Lin, Chuanqi Tan, Chang Zhou, and Jingren Zhou. 2023.
\newblock \# instag: Instruction tagging for analyzing supervised fine-tuning of large language models.
\newblock In \emph{The Twelfth International Conference on Learning Representations}.

\bibitem[{Luo et~al.(2023{\natexlab{a}})Luo, Sun, Xu, Zhao, Lou, Tao, Geng, Lin, Chen, and Zhang}]{luo2023wizardmath}
Haipeng Luo, Qingfeng Sun, Can Xu, Pu~Zhao, Jianguang Lou, Chongyang Tao, Xiubo Geng, Qingwei Lin, Shifeng Chen, and Dongmei Zhang. 2023{\natexlab{a}}.
\newblock Wizardmath: Empowering mathematical reasoning for large language models via reinforced evol-instruct.
\newblock \emph{arXiv preprint arXiv:2308.09583}.

\bibitem[{Luo et~al.(2023{\natexlab{b}})Luo, Xu, Zhao, Sun, Geng, Hu, Tao, Ma, Lin, and Jiang}]{luo2023wizardcoder}
Ziyang Luo, Can Xu, Pu~Zhao, Qingfeng Sun, Xiubo Geng, Wenxiang Hu, Chongyang Tao, Jing Ma, Qingwei Lin, and Daxin Jiang. 2023{\natexlab{b}}.
\newblock Wizardcoder: Empowering code large language models with evol-instruct.
\newblock \emph{arXiv preprint arXiv:2306.08568}.

\bibitem[{Ouyang et~al.(2022)Ouyang, Wu, Jiang, Almeida, Wainwright, Mishkin, Zhang, Agarwal, Slama, Ray et~al.}]{ouyang2022training}
Long Ouyang, Jeffrey Wu, Xu~Jiang, Diogo Almeida, Carroll Wainwright, Pamela Mishkin, Chong Zhang, Sandhini Agarwal, Katarina Slama, Alex Ray, et~al. 2022.
\newblock Training language models to follow instructions with human feedback.
\newblock \emph{Advances in neural information processing systems}, 35:27730--27744.

\bibitem[{Tan et~al.(2024)Tan, Li, Wang, Beigi, Jiang, Bhattacharjee, Karami, Li, Cheng, and Liu}]{tan2024large}
Zhen Tan, Dawei Li, Song Wang, Alimohammad Beigi, Bohan Jiang, Amrita Bhattacharjee, Mansooreh Karami, Jundong Li, Lu~Cheng, and Huan Liu. 2024.
\newblock Large language models for data annotation and synthesis: A survey.
\newblock In \emph{Proceedings of the 2024 Conference on Empirical Methods in Natural Language Processing}, pages 930--957.

\bibitem[{Taori et~al.(2023)Taori, Gulrajani, Zhang, Dubois, Li, Guestrin, Liang, and Hashimoto}]{taori2023alpaca}
Rohan Taori, Ishaan Gulrajani, Tianyi Zhang, Yann Dubois, Xuechen Li, Carlos Guestrin, Percy Liang, and Tatsunori~B Hashimoto. 2023.
\newblock Alpaca: A strong, replicable instruction-following model.
\newblock \emph{Stanford Center for Research on Foundation Models. https://crfm. stanford. edu/2023/03/13/alpaca. html}, 3(6):7.

\bibitem[{Wang et~al.(2022)Wang, Kordi, Mishra, Liu, Smith, Khashabi, and Hajishirzi}]{wang2022self}
Yizhong Wang, Yeganeh Kordi, Swaroop Mishra, Alisa Liu, Noah~A Smith, Daniel Khashabi, and Hannaneh Hajishirzi. 2022.
\newblock Self-instruct: Aligning language models with self-generated instructions.
\newblock \emph{arXiv preprint arXiv:2212.10560}.

\bibitem[{Wang et~al.(2023)Wang, Zhong, Li, Mi, Zeng, Huang, Shang, Jiang, and Liu}]{wang2023aligning}
Yufei Wang, Wanjun Zhong, Liangyou Li, Fei Mi, Xingshan Zeng, Wenyong Huang, Lifeng Shang, Xin Jiang, and Qun Liu. 2023.
\newblock Aligning large language models with human: A survey.
\newblock \emph{arXiv preprint arXiv:2307.12966}.

\bibitem[{Xu et~al.(2023)Xu, Sun, Zheng, Geng, Zhao, Feng, Tao, and Jiang}]{xu2023wizardlm}
Can Xu, Qingfeng Sun, Kai Zheng, Xiubo Geng, Pu~Zhao, Jiazhan Feng, Chongyang Tao, and Daxin Jiang. 2023.
\newblock Wizardlm: Empowering large language models to follow complex instructions.
\newblock \emph{arXiv preprint arXiv:2304.12244}.

\bibitem[{Xu et~al.(2024)Xu, Jiang, Niu, Deng, Poovendran, Choi, and Lin}]{xu2024magpie}
Zhangchen Xu, Fengqing Jiang, Luyao Niu, Yuntian Deng, Radha Poovendran, Yejin Choi, and Bill~Yuchen Lin. 2024.
\newblock Magpie: Alignment data synthesis from scratch by prompting aligned llms with nothing.
\newblock \emph{arXiv preprint arXiv:2406.08464}.

\bibitem[{Yang et~al.(2024)Yang, Yang, Zhang, Hui, Zheng, Yu, Li, Liu, Huang, Wei et~al.}]{yang2024qwen2}
An~Yang, Baosong Yang, Beichen Zhang, Binyuan Hui, Bo~Zheng, Bowen Yu, Chengyuan Li, Dayiheng Liu, Fei Huang, Haoran Wei, et~al. 2024.
\newblock Qwen2. 5 technical report.
\newblock \emph{arXiv preprint arXiv:2412.15115}.

\bibitem[{Zeng et~al.(2024)Zeng, Xu, Zhao, Lou, and Chen}]{zeng2024automatic}
Weihao Zeng, Can Xu, Yingxiu Zhao, Jian-Guang Lou, and Weizhu Chen. 2024.
\newblock Automatic instruction evolving for large language models.
\newblock \emph{arXiv preprint arXiv:2406.00770}.

\bibitem[{Zheng et~al.(2023)Zheng, Chiang, Sheng, Zhuang, Wu, Zhuang, Lin, Li, Li, Xing et~al.}]{zheng2023judging}
Lianmin Zheng, Wei-Lin Chiang, Ying Sheng, Siyuan Zhuang, Zhanghao Wu, Yonghao Zhuang, Zi~Lin, Zhuohan Li, Dacheng Li, Eric Xing, et~al. 2023.
\newblock Judging llm-as-a-judge with mt-bench and chatbot arena.
\newblock \emph{Advances in Neural Information Processing Systems}, 36:46595--46623.

\bibitem[{Zheng et~al.(2024)Zheng, Chiang, Sheng, Zhuang, Wu, Zhuang, Lin, Li, Li, Xing et~al.}]{zheng2024judging}
Lianmin Zheng, Wei-Lin Chiang, Ying Sheng, Siyuan Zhuang, Zhanghao Wu, Yonghao Zhuang, Zi~Lin, Zhuohan Li, Dacheng Li, Eric Xing, et~al. 2024.
\newblock Judging llm-as-a-judge with mt-bench and chatbot arena.
\newblock \emph{Advances in Neural Information Processing Systems}, 36.

\bibitem[{Zhou et~al.(2024)Zhou, Liu, Xu, Iyer, Sun, Mao, Ma, Efrat, Yu, Yu et~al.}]{zhou2024lima}
Chunting Zhou, Pengfei Liu, Puxin Xu, Srinivasan Iyer, Jiao Sun, Yuning Mao, Xuezhe Ma, Avia Efrat, Ping Yu, Lili Yu, et~al. 2024.
\newblock Lima: Less is more for alignment.
\newblock \emph{Advances in Neural Information Processing Systems}, 36.

\bibitem[{Zhou et~al.(2023)Zhou, Lu, Mishra, Brahma, Basu, Luan, Zhou, and Hou}]{zhou2023instruction}
Jeffrey Zhou, Tianjian Lu, Swaroop Mishra, Siddhartha Brahma, Sujoy Basu, Yi~Luan, Denny Zhou, and Le~Hou. 2023.
\newblock Instruction-following evaluation for large language models.
\newblock \emph{arXiv preprint arXiv:2311.07911}.

\end{thebibliography}

% \clearpage

\appendix
\section{Training Hyper-parameter}
\label{app:hyper}
% As shown in Table \ref{tab:hyper},
% The hyper-parameters employed in our experiments are presented in Table \ref{tab:hyper}.

\begin{table}[h]
\centering
\begin{tabular}{lc}
\toprule
\textbf{Hyperparameter} & \textbf{Value} \\
\midrule
\multicolumn{2}{l}{\textbf{General}} \\
Number of Epochs & 3 \\
Number of Devices & 8 \\
Per-device Batch Size & 4 \\
% Gradient Accumulation Steps & 4 \\
Learning Rate & $2 \times 10^{-5}$ \\
Learning Rate Scheduler & WarmupDecayLR \\
Warmup step & 15 \\
Max Sequence Length & 2048 \\
\midrule
\multicolumn{2}{l}{\textbf{Code Dataset}} \\
Gradient Accumulation Steps & 16 \\
\midrule
\multicolumn{2}{l}{\textbf{Math Dataset}} \\
Gradient Accumulation Steps & 8 \\
\bottomrule
\end{tabular}
\caption{Hyperparameters and their values}
\label{tab:hyper}
\end{table}

\section{Prompt Details}
\label{app:prompt}
We have designed a simple and effective prompt
to guide LLM for tag pool construction (Table \ref{tab:prompt_tag}) and tag sampling evolution
(Table \ref{tab:prompt_evolution}).

\begin{table*}[h]
    \centering
    \begin{tabular}{p{0.9\textwidth}} % 使用 p{宽度} 格式
    \toprule
    \textbf{Prompt for Tagging} \\
    \midrule
    You are a tagging system that provides useful tags for task intentions to distinguish tasks for a helpful AI assistant.\\
    \#Task  \textcolor{blue}{\{Instruction\}} \\
    Please follow the steps below to assign tags to the given task.\\
    \textbf{Step 1:} Consider from which aspects that tags can be assigned to cover main features of the task and provide a brief explanation. 
    Aspects need to be summarizing, such as 'Required skill'. etc. 
    Please summarize this task with as few aspects as possible. \\
    \textbf{Step 2:} Based on the \#Aspect List\# obtained in Step 1, assign core tags to the given task from each aspect.
    \bigskip
    
    \textbf{Please reply strictly in the following format:}\\
    Step 1  \#Aspect List and Explanation\#: \\
    Step 2 \#Aspect2Tags\#: \\
        
        \#Aspect2Tags\# \\
        \{"xxx":[tag1, tag2, ...], "xxx":[tag1, tag2, ...], ...\} \\
        where xxx means aspect you get in step 1. \\
    \bottomrule
    \end{tabular}
    \caption{Prompt for Tagging. \textcolor{blue}{Instruction} stands for the instruction used to extract the tags.}
    \label{tab:prompt_tag}

\end{table*}
\begin{table*}[h]
    \centering
    \begin{tabular}{p{0.9\textwidth}} % 使用 p{宽度} 格式
    \toprule
    \textbf{Prompt for Evolution} \\
    \midrule
    You are an Instruction Rewriter that rewrites the given \#Instruction\# into a more challenging version based on the given tags.

Here is the \#Instruction\#: \textcolor{blue}{\{Instruction\}}

Here is the \#Tag List\#:
\textcolor{blue}{\{TagList\}}

Please follow the steps below to rewrite the given \#Instruction\# into a more intricate and demanding version.

\textbf{Step 1:} Carefully read the \#Instruction\# and \#Tag List\#. Select a subset from the \#Tag List\# that will allow the \#Instruction\# to evolve and become more complex. The chosen subset should provide richer, more nuanced information that enhances the original prompt, ultimately increasing its difficulty and quality. Aim to challenge advanced AI assistants like ChatGPT and GPT-4. The subset should contain \textcolor{blue}{\{Budget\}} tags and must exclude any tags already present in the \#Instruction\#.

\textbf{Step 2:} Develop a comprehensive plan based on the \#Tag subset\# generated in Step 1 to make the \#Instruction\# more challenging. This plan should focus on seamlessly integrating multiple tags from the \#Tag subset\# into the original \#Instruction\#. 

\textbf{Step 3:} Execute the plan step by step and provide the \#Rewritten Instruction\#. The \#Rewritten Instruction\# may only add between 10 and 20*\textcolor{blue}{\{Budget\}} words to the original \#Instruction\#.

\textbf{Step 4:} Thoroughly review the \#Rewritten Instruction\# to identify any inconsistencies. Ensure that the \#Rewritten Instruction\# is solely a more challenging version of the original \#Instruction\#.Provide only the \#Final Rewritten Instruction\# without any additional explanation.
\bigskip

\textbf{Please reply strictly in the following format:}\\
Step 1 \#Tag subset\#:\\
Step 2 \#Plan\#:\\
Step 3 \#Rewritten Instruction\#:\\
Step 4 \#Finally Rewritten Instruction\#:\\
    \bottomrule
    \end{tabular}
    \caption{Prompt for Evolution. \textcolor{blue}{Instruction} represents based instruction for evolution; \textcolor{blue}{TagList} represents tags batch for selection; \textcolor{blue}{Budget} represents the number of tags selected each time.}
    \label{tab:prompt_evolution}

\end{table*}

\section{Case Study}
We analyze the examples in code (Table \ref{tab:code_case}) and math (Table \ref{tab:math_case}) for three methods, including Evol-Instrut, Auto Evol-Instrut and Tag-Evol (Ours).
It can be found that the other two methods yield similar outcomes over three rounds of evolution. However, our method stands out in terms of diversity. Moreover, by sampling a greater number of tags in just one round, we are able to generate instructions that are more challenging compared to those produced by the other methods.

\begin{table*}[h]
    \centering
    \begin{tabular}{p{0.9\textwidth}} % 使用 p{宽度} 格式
    \toprule
    \textbf{Original}:Reverse the string given in the input
Reverse the string given in the input \\
    \midrule
\textbf{Evol-Instruct}

\textbf{Round1:}Reverse the string given in the input, and explain the algorithmic steps and time complexity involved in the process.

\textbf{Round2:}Reverse the alphanumeric string provided in the input, detailing the recursive algorithmic steps and analyzing the time complexity in Big O notation.

\textbf{Round3:}Reverse the hexadecimal string provided in the input, detailing the recursive algorithmic steps and analyzing the time complexity in Big O notation, considering the impact of string length on performance.
\\
\midrule

\textbf{Auto Evol-Instruct}

\textbf{Round1:}Given an input string, reverse it only if its length exceeds 5 characters; otherwise, return the original string.

\textbf{Round2:}Given an input string, reverse it and convert to uppercase only if its length exceeds 5 characters and it contains only alphabetic characters; otherwise, return the original string or a message indicating non-alphabetic characters.

\textbf{Round3:}Given an input string, reverse it and convert to uppercase only if its length exceeds 5 characters, it contains only alphabetic characters, more than 2 vowels, and the substring "abc"; otherwise, return the original string or a message indicating the specific reason for non-conversion.
\\
\midrule
\textbf{Tag-Instruct}

\textbf{Tag 1}:Reverse the string given in the input and multiply its length by 2.

\textbf{Tag subset}:['basic math calculations']

\textbf{Tag 3}:Reverse the string given in the input, ensuring it contains both letters and numbers. Process the form to validate the input, and if the string length is even, append the sum of the digits in the string to the reversed output.

\textbf{Tag subset}:['form processing', 'letters and numbers', 'basic math calculations']

\textbf{Tag 5}:Reverse the string given in the input, but first, process the input form to ensure it contains both letters and numbers. For each letter, convert it to its corresponding ASCII value, and for each number, double it. Then, calculate the position of the third character in the modified string and append this position to the end of the reversed string. Provide a coding example to demonstrate the process.

\textbf{Tag subset}:['form processing', 'letters and numbers', 'coding example', 'basic math calculations', 'third character']\\
    \bottomrule
    \end{tabular}
    \caption{Multi-round case on code domain for Evol-Instruct, Auto Evol-Instruct and Tag-Instruct}
    \label{tab:code_case}

\end{table*}

\begin{table*}[h]
    \centering
    \begin{tabular}{p{0.9\textwidth}} % 使用 p{宽度} 格式
    \toprule
\textbf{Original}: Expand $(x-2)(x+2)(x^2+4)$. \\
    \midrule
\textbf{Evol-Instruct}

\textbf{Round1:} Expand $(x-2)(x+2)(x^2+4)$ and explain the significance of the resulting polynomial in the context of algebraic identities.

\textbf{Round2:} Expand $(x-2)(x+2)(x^2+4)$, explain the significance of the resulting polynomial in the context of algebraic identities, and discuss its applications in solving higher-degree polynomial equations.

\textbf{Round3:} Expand $(x-2)(x+2)(x^2+4)$, explain how the resulting polynomial exemplifies the sum and difference of squares identity, and discuss its applications in factoring and solving quartic equations.
\\
\midrule

\textbf{Auto Evol-Instruct}

\textbf{Round1:} Expand \((x-a)(x+2+ax)(x^2+4a)\) where \(a\) is a non-zero integer.

\textbf{Round2:} Expand \((x-a)(x+2+b+ax)(x^2+4a)\) where \(a = x + 1\), \(b\) is a positive integer, and \(a\) is a non-zero integer.

\textbf{Round3:} Expand \((x-a)(x+2+b+ax)(x^2+4a)\) where \(a = x + 1\), \(b = 2x + 1\), \(c = x^2 + b\), \(a\) is a non-zero integer, \(b\) is a positive integer, and \(c > x\).
\\
\midrule

\textbf{Tag-Instruct}

\textbf{Tag 1}: Expand $(x-2)(x+2)(x^2+4)$, given that $x$ is an integer such that $x > 0$.

\textbf{Tag subset}:['conditions on variables']

\textbf{Tag 3}: Expand $(x-2)(x+2)(x^2+4)$ for $x > 0$, then use the result to find the perimeter of a square whose side length is the square root of the expanded expression evaluated at $x = 3$.

\textbf{Tag subset}:['conditions on variables', 'sequential operations', 'polygon perimeter']

\textbf{Tag 5}: Expand $(x-2)(x+2)(x^2+4)$, where $x$ is an integer such that $-3 \leq x \leq 3$. Factor the expanded polynomial, if possible, and then evaluate $\sin(\pi \cdot \text{expanded polynomial})$. Find the sum of all integer values of $x$ for which the result is an integer.

\textbf{Tag subset}:['conditions on variables', 'factor', 'evaluating trigonometric functions', 'sequential operations', 'integer sum problem']
\\
    \bottomrule
    \end{tabular}
    \caption{Multi-round case on math domain for Evol-Instruct, Auto Evol-Instruct and Tag-Instruct}
    \label{tab:math_case}
\end{table*}

\end{document}